\documentclass[10pt,twocolumn,letterpaper]{article}

\usepackage{wacv}
\usepackage{times}
\usepackage{epsfig}
\usepackage{graphicx}
\usepackage{amsmath}
\usepackage{amssymb}
\usepackage{subcaption}
\usepackage{multirow}
\usepackage[accsupp]{axessibility}  

\def\wacvPaperID{669} 

\wacvfinalcopy 

\ifwacvfinal
\fi

\ifwacvfinal
\usepackage[breaklinks=true,bookmarks=false]{hyperref}
\else
\usepackage[pagebackref=true,breaklinks=true,colorlinks,bookmarks=false]{hyperref}
\fi

\begin{document}

\title{Disentangled Representation with Dual-stage \\
Feature Learning for Face Anti-spoofing}
\author{Yu-Chun Wang$^{*}$, Chien-Yi Wang$^{\dag}$, Shang-Hong Lai$^{*\dag}$\\
$^{*}$National Tsing Hua University, $^{\dag}$Microsoft AI R\&D Center, Taiwan\\
{\tt\small yuchun@gapp.nthu.edu.tw chiwa@microsoft.com shlai@microsoft.com}

}
\maketitle

\ifwacvfinal
\thispagestyle{empty}
\fi

\begin{abstract}
As face recognition is widely used in diverse security-critical applications, the study of face anti-spoofing (FAS) has attracted more and more attention. Several FAS methods have achieved promising performance if the attack types in the testing data are included in the training data, while the performance significantly degrades for unseen attack types. It is essential to learn more generalized and discriminative features to prevent overfitting to pre-defined spoof attack types. This paper proposes a novel dual-stage disentangled representation learning method that can efficiently untangle spoof-related features from irrelevant ones. Unlike previous FAS disentanglement works with one-stage architecture, we found that the dual-stage training design can improve the training stability and effectively encode the features to detect unseen attack types. Our experiments show that the proposed method provides superior accuracy than the state-of-the-art methods on several cross-type FAS benchmarks. 
\end{abstract}


\section{Introduction}

Face recognition technologies have been frequently applied in our daily life in recent years. The associated facial recognition security issues become critical because hackers may easily break into the face verification system by different presentation attacks.

\begin{figure}[t]
\begin{center}

   \includegraphics[width=1.1\linewidth]{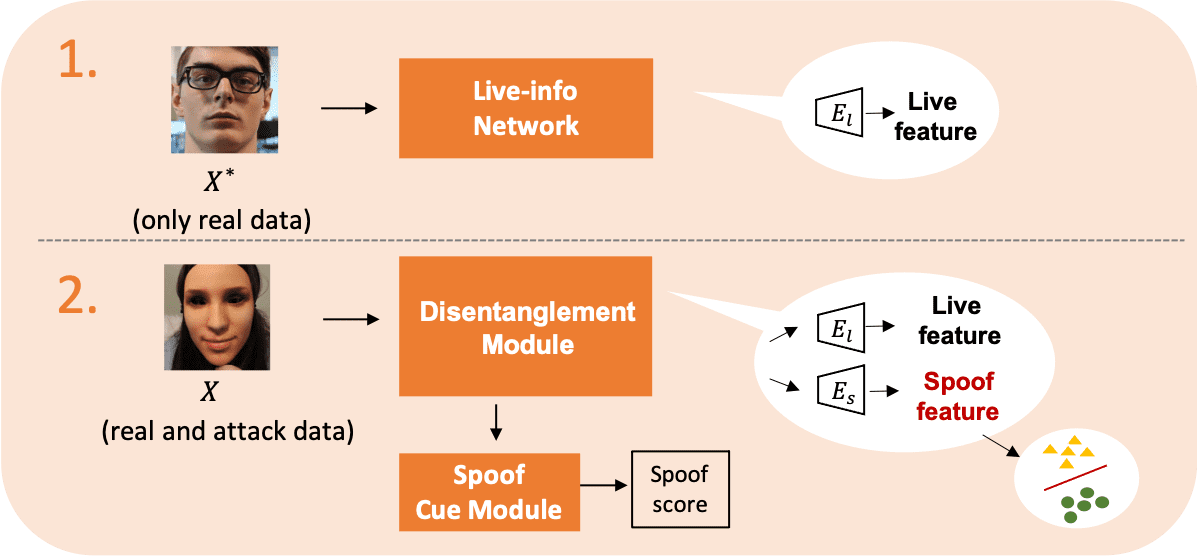}
\end{center}
   \caption{The pipeline of the proposed Dual-stage Feature Learning FAS system. The live-info network concentrates on learning live features in $1^{st}$ stage. In the $2^{nd}$ stage, the disentanglement module adopts the pre-trained Live-encoder in the  $1^{st}$ stage as a fixed live feature extractor and further disentangle spoof features and live features.}
\label{fig:pipeline}
\end{figure}

\begin{figure*}[h]
\begin{center}

  \begin{subfigure}{\textwidth}
    \centering
    \includegraphics[width=0.7\linewidth]{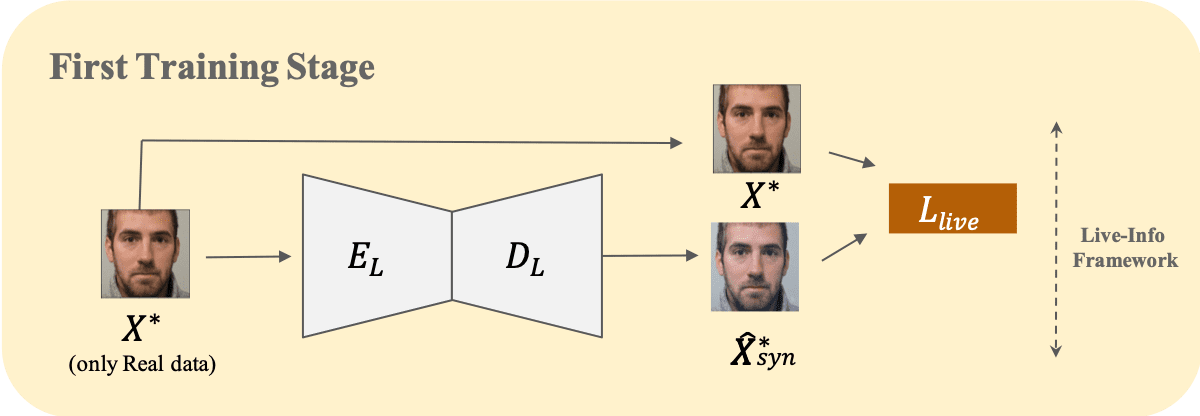}
    \caption{First training stage.}
    \label{fig:Stage1}
  \end{subfigure}
  \hfill

  \vskip\baselineskip
  \begin{subfigure}{\textwidth}   
    \centering 
    \includegraphics[width=0.9\linewidth]{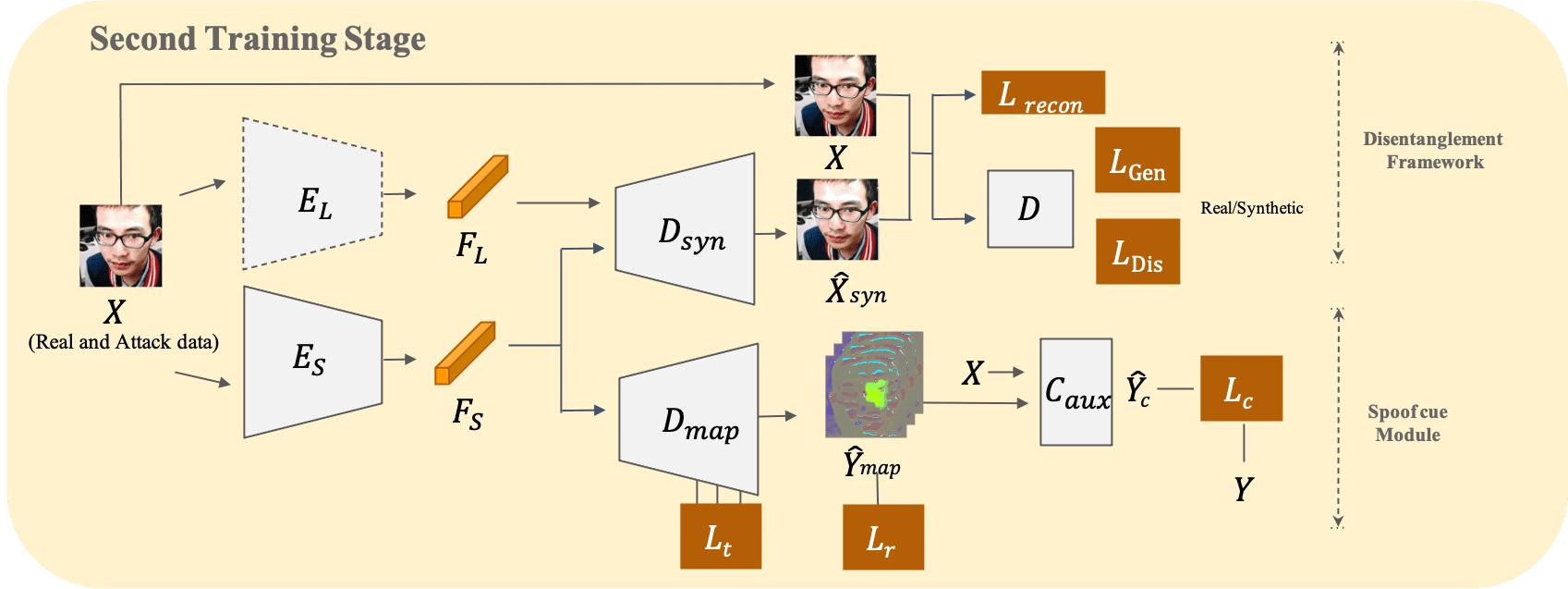}
    \caption{Second training stage.}
    \label{fig:Stage2}
  \end{subfigure}
  \hfill

\end{center}
   \caption{\textbf{Overview of the proposed dual-stage feature learning FAS system:} In the first stage$(a)$, live-info network focuses on learning live features. In the second stage $(b)$, we take pre-trained encoder $E_{L}$ as live feature fixed extractor. By reconstructing original data, disentanglement module can concentrate on extracting disentangled spoof features.}
\label{fig:Stage}
\end{figure*}

The presentation attack types are diverse, which vary from typical printed facial photos and video replays to more high-priced 3D masks.
Attackers can also utilize spoof attacks that the system designers are unaware of; we call them unknown spoof attack types.
As the increasing growth of presentation attacks, face anti-spoofing systems have high probability of encountering unseen presentation instruments.

CNN-based methods can detect pre-defined attack types well, while the generalization ability of these methods is limited for the seen attack types in the training data. The model trained with the existing spoof types would easily fail to detect unseen types of spoof attacks. Some CNN-based methods employ auxiliary information such as heart signal and facial depth to guide the network to distinguish real humans and spoof attacks~\cite{jourabloo2018face, liu2018learning, yu2020face, lin2019face}. Although auxiliary supervision helps extract more discriminative features, the performances of FAS models trained with auxiliary supervisions rely on the accuracy of the additional tasks. Besides, existing auxiliary information may not be suited for all attack types, especially unseen attacks.

Detection of unseen attacks is a serious problem. It is impossible to define all auxiliary distinguishing information to guide the model training for detecting spoofs. The most crucial part lies in the model learning the essence of distinguishing real faces from spoof faces to avoid the network overfitting to training data. Disentangled representation learning is an effective approach to separate the model's latent representation into interpretable parts, further improving the model's robustness~\cite{zhang2020face, liu2020disentangling}. These works proposed different strategies to enforce the model learning disentangled features in an end-to-end architecture, and they use adversarial learning to facilitate disentangled representation.
The stability of the generator and discriminator would largely influence the disentanglement representation effect.
Furthermore, it is difficult to guarantee that the models can precisely extract disentangled representation for live and spoof components at the same time.

In light of these problems, we aim to improve the unknown attack detection accuracy with a two-stage disentanglement representation architecture to guarantee that the model extracts more stable disentangled representations. While many previous works leverage auxiliary supervision during training, we build our model without supervision from auxiliary tasks, whose ground truths may not be accurate or suitable for the target task.


The main contributions of our paper are summarized as follows:

\begin{enumerate}
\item  We propose a novel dual-stage feature disentangled representation learning method for the FAS problem. Our disentangled spoof features provide powerful generalization ability when facing unknown attacks.

\item We propose to regress spoof maps through our disentangled spoof feature. This method can further improve our model generalization ability against unknown attacks. 

\item Our experiments show that our proposed method outperforms the state-of-the-art methods on several  cross-type FAS protocols. 
\end{enumerate}

\section{Related Work}

\subsection{Face Anti-Spoofing}

Traditional FAS methods utilize classical local descriptors such as LBP~\cite{maatta2011face}, HOG~\cite{komulainen2013context}, and SIFT~\cite{patel2016secure} to extract hand-crafted features as the facial images textures. After that, they use classic classifiers like SVM and LDA to make a binary decision. Other methods focus on motion-based face anti-spoofing approaches; they use motion cues like eye-blinking ~\cite{pan2007eyeblink}, mouth movement ~\cite{de2014face}, and head movement ~\cite{kollreider2007real}  to detect print-out attacks. However, motion-based methods fail against 3D mask attacks since some masks have well-preserved motion cues like eyes and mouth ~\cite{liu20163d}.

With the development of CNN, many researchers introduce the CNN model as a feature extractor and formulate FAS as a binary classification problem~\cite{george2019deep, yu2020multi,jourabloo2018face}. In order to learn discriminative features efficiently, they adopt auxiliary supervisions like depth~\cite{liu2018learning}, reflection~\cite{kim2019basn}, and rRPG~\cite{liu2018learning,liu2018remote}. Despite applying task-oriented knowledge, deep learning based approaches have limited generalization capability. They tend to overfit to pre-defined spoof attack types in the training data. When the model encounters unseen spoof attacks, the FAS performance would drop drastically.

\subsection{Disentangled representation learning}
Disentangled representation is an unsupervised learning method that can untangle data into interpretable components. 
The well-disentangled spoof-related features could help us enhance the robustness of the FAS model. GAN and VAE architectures have the powerful capability to generate new data. Therefore, researchers utilize them to learn disentangled representations\cite{xiao2018elegant, huang2018multimodal}. \cite{liu2020disentangling} disentangles spoof traces into a hierarchical combination and further synthesizes new data to enhance training. \cite{zhang2020face} separates representation into liveness and content features, and they combine depth and texture information to boost training. These FAS works focus on solving common FAS problems; therefore, they utilize end-to-end strategies to limit their model disentangling spoof features. However, to detect unknown spoofing attacks, we require more generalized spoofing features. Different from previous work, We propose a dual-stage feature-disentangled representation learning method to extract more discriminative spoofing features.  

\section{Proposed method}
We assume that every spoof face is composed of its real counterpart and spoof-related information. Accordingly, we split the latent representation of facial data into live feature and spoof feature. Live features correlate with the information that spoof images share with their real counterparts, such as identification and background. Spoof features are the spoof-related information that differentiates between attack-face and real-face images.

To disentangle the live feature and spoof feature efficiently, we propose a dual-stage feature learning method to disentangle face representation.
Our training process has two stages. The first stage is based on the live-info network, and the second stage consists of the disentanglement module and Spoof Cue module. The whole learning process is illustrated in Figure~\ref{fig:Stage}.

\subsection{Live-info Network}

In the first training stage, we only utilize real images as training data to train the live-info network. Figure \ref{fig:Stage1} shows the network architecture of our live-info network.
The Live-encoder $E_{L}$ is a one-class auto-encoder. We use loss $L_{Live}$ to constrain the model to extract live features and reconstruct real face images accurately.
Because the model only learns information from real samples, it can extract the universal factors of variation from real data.
We regard all the features extracted by the Live-encoder $E_{L}$ as live features. 

Let $X^{*}$ denote the real training data. The live-info network can generate output $\hat{X}_{syn}^{*} = D_{L}(E_{L}(X^{*}))$. The loss function $L_{Live}$ is given by
\begin{equation}
\begin{split}
L_{Live} = \mathbb{E}_{x\sim P_{X}}[
||{{x}^{*}-\hat{x}^{*}_{syn}}\|_2^2]
\end{split}
\label{L_live}
\end{equation}

\subsection{Disentanglement Module}

As shown in Figure \ref{fig:Stage2}, we utilize both real and attack samples as training data in the second stage. Two-branch encoders $E_{L}$ and $E_{S}$ are used to extract the live features $F_{L}$ and spoof features $F_{S}$, respectively.
The Live-encoder $E_{L}$ loads the pre-trained model weights in the first stage as a live feature fixed extractor. Since the Live-encoder $E_{L}$ had not learned any information about spoof images, it can focus on extracting spoof-invariant features. In order to reconstruct complete input data, the Spoof-encoder $E_{S}$ concentrates on extracting the information that differentiates attack and real data. We employ the loss $L_{recon}$ and adversarial loss to facilitate the outputs $\hat{X}_{syn}$ similar to the original images $X$. We take the generated data $\hat{X}_{syn}$ and original data $X$ as the input of the discriminator $D$, and it will determine if the input samples are real or synthetic.

Let $(X, Y)$ denote the input data and the corresponding labels from all the training data.  The adversarial loss is formed by 
\begin{equation}
\begin{split}
L_{Gen} = \mathbb{E}_{x\sim P_{X}} [(D(\hat{x}_{syn})-1)^2] 
\end{split}
\end{equation}
and
\begin{equation}
\begin{split}
L_{Dis} =  \mathbb{E}_{x\sim P_{X}} [(D(x) - 1)^2]+\\
\mathbb{E}_{x\sim P_{X}}[(D(\hat{x}_{syn}))^2]\\
\end{split}
\end{equation}

The reconstruction loss $L_{recon}$ is given by 
\begin{equation}
\begin{split}
L_{recon} = \mathbb{E}_{x\sim P_{X}}[
||{{x}-\hat{x}_{syn}}\|_2^2]
\end{split}
\end{equation}

\subsection{Spoof Cue Module}

We use the disentangled spoof feature to decode spoof maps in an unsupervised way. It can avoid our approach from overfitting to known attacks in the training data. We modify the method proposed in \cite{feng2020learning} as our Spoof Cue Module. They adopt a U-Net\cite{ronneberger2015u} to generate spoof cue maps.
In our Spoof Cue Module, we utilize disentangled spoof features $F_{S}$ in the disentanglement module as input and adopt a decoder $D_{map}$ to generate spoof maps.

 The loss function $L_{t}$ is used to facilitate contracting the spoof features $F_{S}$ of real face images and separating the spoof features between real and spoof data. The loss $L_{t}$ is applied in the last three layers of decoder $D_{map}$.
Let {$f_{i}^{a}$, $f_{i}^{p}$, $f_{i}^{n}$} denote the  feature vectors of anchor (real), positive (real), and negative (attack) samples, respectively, for the $i$-th triplet. We always pick live sample as the anchor. The loss  function $L_{t}$ consists of two parts, i.e., $L_{t}^{normal}$ and $L_{t}^{hard}$.

The loss $L_{t}^{normal}$ counts normal triplet loss for all valid triplets and averages those whose values are positive.
N denotes the number of triplets. $\alpha$ is a pre-defined margin constant.

\begin{equation}
\begin{split}
L_{t}^{normal} = \dfrac{1}{N}\sum ^{N}_{i=1}\max(\| f_{i}^{a}-f_{i}^{p}\| _{2}^{2}-  \| f_{i}^{a}-f_{i}^{n}\|_{2}^{2}+\alpha,0)
\end{split}
\end{equation}

The loss $L_{t}^{hard}$ is a hard-triplet loss.
For each anchor $f_{i}^{a}$, we take the pair ($f_{i}^{a}$,$f_{j}^{p}$)  which has the  largest Euclidean distance as the hard positive pair. The positive sample index $j$ is selected from $L$ and $j \neq i$, where $L$ denotes the set of all live samples in the current batch. For each anchor $f_{i}^{a}$, we find the pair ($f_{i}^{a}$,$f_{k}^{n}$)  with the smallest Euclidean distance as the hard negative pair. The spoofing sample index $k$ is selected from $N$, where $N$ denotes the set of all spoofing samples in the current batch. T denotes the number of triplets. $m$ is the pre-defined margin constants. For each triplet ($f_{i}^{a}$,$f_{i}^{p}$,$f_{i}^{n}$), we compute each hard triplet loss, and then average them.
\begin{equation}
\begin{split}
L_{t}^{hard} = \dfrac{1}{T}\sum ^{T}_{i=1}\max(\max_{j \in L} \| f_{i}^{a}-f_{j}^{p}\|_{2}^{2}- \\ 
\min_{k \in N} \| f_{i}^{a}-f_{k}^{n}\|_{2}^{2}+m,0)
\end{split}
\end{equation}
where $m$ is a pre-defined margin.

The loss function $L_{t}$ is the sum of the above two loss functions, i.e.
\begin{equation}
\begin{split}
L_{t} = L_{t}^{normal} + L_{t}^{hard} 
\end{split}
\end{equation}

We need to generate the discriminative spoof map $\hat{y}_{map}$, which represents the difference between real data and attack data.
For this purpose, we use $L_{r}$ and $L_{c}$ to facilitate training the decoder $D_{map}$.

We use loss $L_{r}$ to minimize the spoof map of real data while without setting constraints for spoofing data. $x \in P_{live}$ denotes the real input data.

\begin{equation}
\begin{split}
L_{r}= \mathbb{E}_{x\sim P_{live}}[\left \|\hat{y}_{map} \right \|_{1}] 
\end{split}
\end{equation}

$C_{aux}$ is a binary classifier to strengthen the $D_{map}$ decoding spoof map $\hat{y}_{map}$. We overlap spoof map $\hat{y}_{map}$ with original data $X$ as the input of $C_{aux}$. The loss $L_{c}$ for training the classifier $C_{aux}$ is given by
\begin{equation}
L_{c} = \mathbb{E}_{x\sim P_{X},y\sim P_{Y}} -y\cdot log(\hat{y}_{c})
+(1-y)\cdot log(1-\hat{y}_{c}))
\end{equation}
where $y$ denotes the binary label, and $\hat{y}_{c}$ is the prediction of the classifier $C_{aux}$.

\subsection{Training and Testing}
We weight the summation of the loss functions described above as the final loss in the second training stage, given by
\begin{equation}
\begin{split}
L = \lambda _{1}L_{recon} + \lambda _{2}L_{Gen} + \lambda _{3}L_{t} + \lambda _{4}L_{r} + \lambda _{5}L_{c}
\end{split}
\end{equation}
where $\lambda _{1}$, $\lambda _{2}$, $\lambda _{3}$, $\lambda _{4}$ and $\lambda _{5}$ are the weights associated with the loss functions described above, and their values are set to 4, 1, 3, 5, and 5, respectively, in our experiments.

At the testing phase, we calculate the average of the generated spoof map $\hat{y}_{map}$ as the spoof score and further utilize it to make the real/attack decision. The input data with large scores are likely to be a spoof one. 
We only need to use the encoder $E_{s}$ and the decoder $D_{map}$ in the inference stage; therefore, our method can achieve 121.48$\pm$1.4 FPS on GeForce GTX 1080.

\begin{table*}[ht]
\hspace*{-0.8cm}
\resizebox{1.1\textwidth}{!}{\begin{tabular}{c|c|c|c|c|c|c|c|c|c|c|c|c|c|c|c}
\hline
\multirow{2}{*}{Method}    & \multirow{2}{*}{Metrics(\%)} & \multirow{2}{*}{Replay} & \multirow{2}{*}{Print} & \multicolumn{5}{c|}{Mask attacks}         & \multicolumn{3}{c|}{Makeup Attacks} & \multicolumn{3}{c|}{Partial Attacks}   & \multirow{2}{*}{Average} \\ \cline{5-15}
                           &                              &                         &                        & Half & Silicone & Trans. & Paper & Manne. & Obfusc.   & Imperson.   & Cosmetic  & Funnyeye & Paperglasses & Partialpaper &                          \\ \hline\hline
\multirow{4}{*}{\begin{tabular}[c]{@{}c@{}}Auxiliary \cite{liu2018learning}\\  (CVPR 2018)\end{tabular}} & APCER                        & 23.7                    & 7.3                    & 27.7 & 18.2     & 97.8   & 8.3   & 16.2   & 100.0     & 18.0        & 16.3      & 91.8     & 72.2         & 0.4          & 38.3$\pm$37.4               \\ \cline{3-16} 
                           & BPCER                        & 10.1                    & 6.5                    & 10.9 & 11.6     & 6.2    & 7.8   & 9.3    & 11.6      & 9.3         & 7.1       & 6.2      & 8.8          & 10.3         & 8.9$\pm$2.0                 \\ \cline{3-16} 
                           & ACER                         & 16.8                    & 6.9                    & 19.3 & 14.9     & 52.1   & 8.0     & 12.8   & 55.8      & 13.7        & 11.7      & 49.0       & 40.5         & 5.3          & 23.6$\pm$18.5               \\ \cline{3-16} 
                           & EER                          & 14.0                      & 4.3                    & 11.6 & 12.4     & 24.6   & 7.8   & 10.0     & 72.3      & 10.1        & 9.4       & 21.4     & 18.6         & 4.0          & 17.0$\pm$17.7               \\ \hline\hline
\multirow{4}{*}{\begin{tabular}[c]{@{}c@{}}DTN \cite{liu2019deep}\\  (CVPR 2019)\end{tabular}}       & APCER                        & 1.0                     & 0.0                    & 0.7  & 24.5     & 58.6   & 0.5   & 3.8    & 73.2      & 13.2        & 12.4      & 17.0     & 17.0         & 0.2          & 17.1$\pm$23.3               \\ \cline{3-16} 
                           & BPCER                        & 18.6                    & 11.9                   & 29.3 & 12.8     & 13.4   & 8.5   & 23.0   & 11.5      & 9.6         & 16.0      & 21.5     & 22.6         & 16.8         & 16.6$\pm$6.2                \\ \cline{3-16} 
                           & ACER                         & 9.8                     & 6.0                    & 15.0 & 18.7     & 36.0     & 4.5   & 7.7    & 48.1      & 11.4        & 14.2      & 19.3     & 19.8         & 8.5          & 16.8$\pm$11.1               \\ \cline{3-16} 
                           & EER                          & 10.0                    & \textbf{2.1}                    & 14.4 & 18.6     & 26.5   & 5.7   & 9.6    & 50.2      & 10.1        & 13.2      & 19.8     & 20.5         & 8.8          & 16.1$\pm$12.2               \\ \hline\hline
\multirow{4}{*}{\begin{tabular}[c]{@{}c@{}}DST \cite{liu2020disentangling}\\  (ECCV 2020)\end{tabular}}       & APCER                        & 1.6                     & 0.0                    & 0.5  & 7.2      & 9.7    & 0.5   & 0.0    & 96.1      & 0.0         & 21.8      & 14.4     & 6.5          & 0.0          & 12.2$\pm$26.1               \\ \cline{3-16} 
                           & BPCER                        & 14.0                    & 14.6                   & 13.6 & 18.6     & 18.1   & 8.1   & 13.4   & 10.3      & 9.2         & 17.2      & 27.0       & 35.5         & 11.2         & 16.2$\pm$7.6                \\ \cline{3-16} 
                           & ACER                         & \textbf{7.8}                    & 7.3                    & 7.1  & 12.9     & \textbf{13.9}   & 4.3   & 6.7    & 53.2      & 4.6         & 19.5      & 20.7     & 21.0           & 5.6          & 14.2$\pm$13.2               \\ \cline{3-16} 
                           & EER                          & \textbf{7.6}                     & 3.8                    & 8.4  & 13.8     & 14.5   & 5.3   & 4.4    & 35.4      & 0.0         & 19.3      & 21       & 20.8         & 1.6          & 12.0$\pm$10.0               \\ \hline\hline
\multirow{4}{*}{\begin{tabular}[c]{@{}c@{}}HMP \cite{yu2020face}\\  (ECCV 2020)\end{tabular}}       & APCER                        & 12.4                    & 5.2                    & 8.3  & 9.7      & 13.6   & 0.0   & 2.5    & 30.4      & 0.0         & 12.0      & 22.6     & 15.9         & 1.2          & 10.3$\pm$9.1                \\ \cline{3-16} 
                           & BPCER                        & 13.2                    & 6.2                    & 13.1 & 10.8     & 16.3   & 3.9   & 2.3    & 34.1      & 1.6         & 13.9      & 23.2     & 17.1         & 2.3          & 12.2$\pm$9.4                \\ \cline{3-16} 
                           & ACER                         & 12.8                    & 5.7                    & 10.7 & 10.3     & 14.9   & \textbf{1.9}   & \textbf{2.4}    & 32.3      & 0.8         & 12.9      & 22.9     & 16.5         & \textbf{1.7}          & 11.2$\pm$9.2                \\ \cline{3-16} 
                           & EER                          & 13.4                    & 5.2                    & 8.3  & 9.7      & 13.6   & 5.8   & 2.5    & 33.8      & \textbf{0.0}         & 14.0        & 23.3     & 16.6         & \textbf{1.2}          & 11.3$\pm$9.5                \\ \hline\hline
\multirow{4}{*}{\begin{tabular}[c]{@{}c@{}}NAS-FAS \cite{yu2020fas}\\  (TPAMI 2020)\end{tabular}}                     & APCER                        & 12.8                    & 9.0                    & 9.7  & 13.1     & 19.1   & 1.1   & 5.4    & 31.0        & 0.0         & 15.0      & 15.1     & 18.6         & 5.0          & 11.9$\pm$8.4                \\ \cline{3-16} 
                           & BPCER                        & 10.1                    & 6.8                    & 13.1 & 11.1     & 12.5   & 2.8   & 0.0    & 26.1      & 0.8         & 15.3      & 17.8     & 13.5         & 2.3          & 10.2$\pm$7.5                \\ \cline{3-16} 
                           & ACER                         & 9.3                     & 7.9                    & 11.4 & 12.1     & 15.8   & \textbf{1.9}   & 2.7    & 28.5      & \textbf{0.4}         & 15.1      & \textbf{16.5}     & \textbf{16.0}         & 3.8          & 10.9$\pm$7.8                \\ \cline{3-16} 
                           & EER                          & 9.3                     & 6.8                    & 9.7  & 11.1     & \textbf{12.5}   & 2.7   & \textbf{0.0}    & 26.1      & \textbf{0.0}         & 15.0      & \textbf{15.1}     & \textbf{13.4}         & 2.3          & 9.5$\pm$7.4                 \\ \hline\hline
\multirow{4}{*}{\textbf{Ours}}      & APCER                        & 8.16                     & 1.7                    & 0.0  & 3.7      & 11.45  & 0.0   & 0.0    & 17.39     & 1.63        & 8.0       & 41.87    & 27.55        & 6.97         & 9.88$\pm$12.5               \\ \cline{3-16} 
                           & BPCER                        & 7.63                    & 9.16                   & 6.87 & 8.39     & 20.45  & 7.63  & 6.87   & 14.5      & 4.58        & 9.9       & 6.87     & 13.74        & 6.1          & 9.44$\pm$4.36               \\ \cline{3-16} 
                           & ACER                         & 7.89                    & \textbf{5.43}                   & \textbf{3.43} & \textbf{6.05}     & 15.9   & 3.81  & 3.43   & \textbf{15.94}     & 3.1         & \textbf{8.96}      & 24.37    & 20.64        & 6.54         & \textbf{9.65$\pm$7.15}               \\ \cline{3-16} 
                           & EER                          & 7.63                    & 6.87                   & \textbf{5.34} & \textbf{7.63}     & 12.97  & \textbf{0.0}   & 6.1    & \textbf{14.5}      & 3.05        & \textbf{8.39}      & 16.03    & 17.55        & 6.1          & \textbf{8.63$\pm$5.18}               \\ \hline\hline
\end{tabular}
}
\caption{ Comparison of the state-of-the-art methods on the cross-type testing on SiW-M dataset \cite{liu2019deep}.}
\label{siwm}
\end{table*}

\section{Experiments}
\subsection{Databases and Protocols}
We evaluate the proposed method on five popular FAS databases: SiW-M\cite{liu2019deep}, MSU-MFSD\cite{wen2015face}, CASIA-MFSD\cite{zhang2012face}, Replay-Attack\cite{chingovska2012effectiveness}, and SiW\cite{liu2018learning}.
To evaluate the generalization ability of the proposed method in unseen attacks, we test on several cross-type protocols\cite{arashloo2017anomaly} \cite{liu2019deep} like SiW-M\cite{liu2019deep}, MSU-MFSD\cite{wen2015face}, CASIA-MFSD\cite{zhang2012face}, and Replay-Attack\cite{chingovska2012effectiveness}. We also evaluate the intra-type protocol, which adopts identical attack types but with illumination and expression variations, like sub-protocol 1 and 2 in SiW dataset\cite{liu2018learning}.
Moreover, we provide the cross-database result in the supplementary material.

\subsection{Evaluation metrics}
We select five metrics which are commonly used to evaluate the performance in FAS tasks: Equal Error Rate (EER), Attack Presentation Classification Error Rate(APCER), Bona Fide Presentation Classification Error Rate (BPCER), the average of APCER and BPCER, Average Classification Error Rate (ACER) = (APCER+BPCER)/2, and Area Under Curve (AUC).

\subsection{Implementation Details}
We crop the face regions from the datasets as training data by the face detector in Dlib \cite{king2009dlib} for the datasets without providing face location ground truth. Then, the cropped regions are all resized to the size 256x256. We resample the training data to retain the ratio between real images and attack images to 1:1. 
We choose Adam \cite{da2014method} as the optimizer with an initial learning rate of $5e-4$, and the training batch size is 32. The method is implemented by Pytorch. Because our model is a two-stage training procedure, we train the live-info network with 10 epochs in the first stage. In the second stage, we load the pre-trained Live-encoder $E_{L}$ as a fixed live-feature extractor to further train the disentanglement module.
We set the switching frequency of the discriminator and disentanglement module empirically. Our disentanglement module can converge after around ten epochs for all the datasets in our implementation.

\subsection{Testing on SiW-M}

We evaluate the cross-type detection performance on the SiW-M dataset. The SiW-M dataset adopts the leave-one-out testing protocols and has 968 videos of 13 types of spoof attacks. Following the original work \cite{liu2019deep}, we use 12 types of spoof attacks plus 80$\%$ of the real videos to train the model and evaluate the remaining one attack type plus the 20$\%$ of real videos in every sub-protocol. For a fair comparison with other state-of-the-art methods, we adopt the same approach describing in \cite{liu2019deep}. The threshold are decided with the training data and fix it for all testing sub-protocols.
As Table \ref{siwm} shows, our method achieves top performance with average ACER (9.65$\%$) and EER (8.63$\%$) for all the 13 attack types.  Furthermore, our evaluation result has a smaller standard deviation than all the other methods. It shows that our method has a stronger generalization ability when encountering unknown attack types. 

\subsection{Testing on MSU-MFSD, CASIA-MFSD, and Replay-Attack}
We also test the cross-type detection performance on MSU-MFSD, CASIA-MFSD, and Replay-Attack; we follow the unseen attack type protocol proposed in \cite{arashloo2017anomaly}. This protocol also adopts ‘leave one attack type out’ to validate the model robustness against unknown attack types. As illustrated in Table \ref{tb2}, our method exceeds the state-of-the-art results and achieves the best average AUC(98.0$\%$) in this protocol. Specifically, we greatly improve the AUC ($\%$) for the printed spoof type in MSU-MFSD, which type other methods perform worse. It demonstrates that our method is more robust when encountering unseen attacks.

\begin{table*}[ht]
\resizebox{\textwidth}{!}{
\begin{tabular}{c|c|c|c|c|c|c|c|c|c|c}
\hline
\multirow{2}{*}{Method} & \multicolumn{3}{c|}{CASIA-MFSD} & \multicolumn{3}{c|}{Replay-Attack} & \multicolumn{3}{c|}{MSU-MFSD}     & \multirow{2}{*}{Overall} \\ \cline{2-10}
                        & Video   & Cut photo  & Wrapped  & Video  & Digital Photo  & Printed  & Printed & HR video & Mobile Video &                          \\ \hline\hline
SVM+LBP \cite{zinelabidine2017oulunpu}                & 91.94   & 91.7       & 84.47    & 99.08  & 98.17          & 87.28    & 47.68   & 99.5     & 97.61        & 88.55$\pm$16.25             \\ \hline
NN+LBP \cite{xiong2018unknown}                  & 94.16   & 88.39      & 79.85    & 99.75  & 95.17          & 78.86    & 50.57   & 99.93    & 93.54        & 86.69$\pm$16.25             \\ \hline
DTN \cite{liu2019deep}                    & 90      & 97.3       & 97.5     & 99.9   & 99.9           & 99.6     & 81.6    & 99.9     & 97.5         & 95.9$\pm$6.2                \\ \hline
AIM-FAS \cite{qin2020learning}                & 93.6    & 99.7       & 99.1     & 99.8   & 99.9           & 99.8     & 76.3    & 99.9     & 99.1         & 96.4$\pm$7.8                \\ \hline
NAS-FAS \cite{yu2020fas}                & 99.62   & 100        & 100      & 99.99  & 99.89          & 99.98    & 74.62   & 100      & 99.98        & 97.1$\pm$8.9                \\ \hline\hline
\textbf{Ours}                    & 98.05   & 99.36      & 99.83    & 99.59  & 99.96          & 100      & 86.78   & 100      & 98.39        & \textbf{98.0$\pm$4.26 }                \\ \hline\hline
\end{tabular}}
\caption{AUC ($\%$) of the state-of-the-art methods for the cross-type testing on CASIA-MFSD\cite{zhang2012face}, Replay-Attack\cite{chingovska2012effectiveness}, and MSU-MFSD\cite{zhang2012face}.}
\label{tb2}
\end{table*}

\begin{figure*}[ht]
\centering
    \hspace*{-0.6cm}
   \includegraphics[width=1\linewidth]{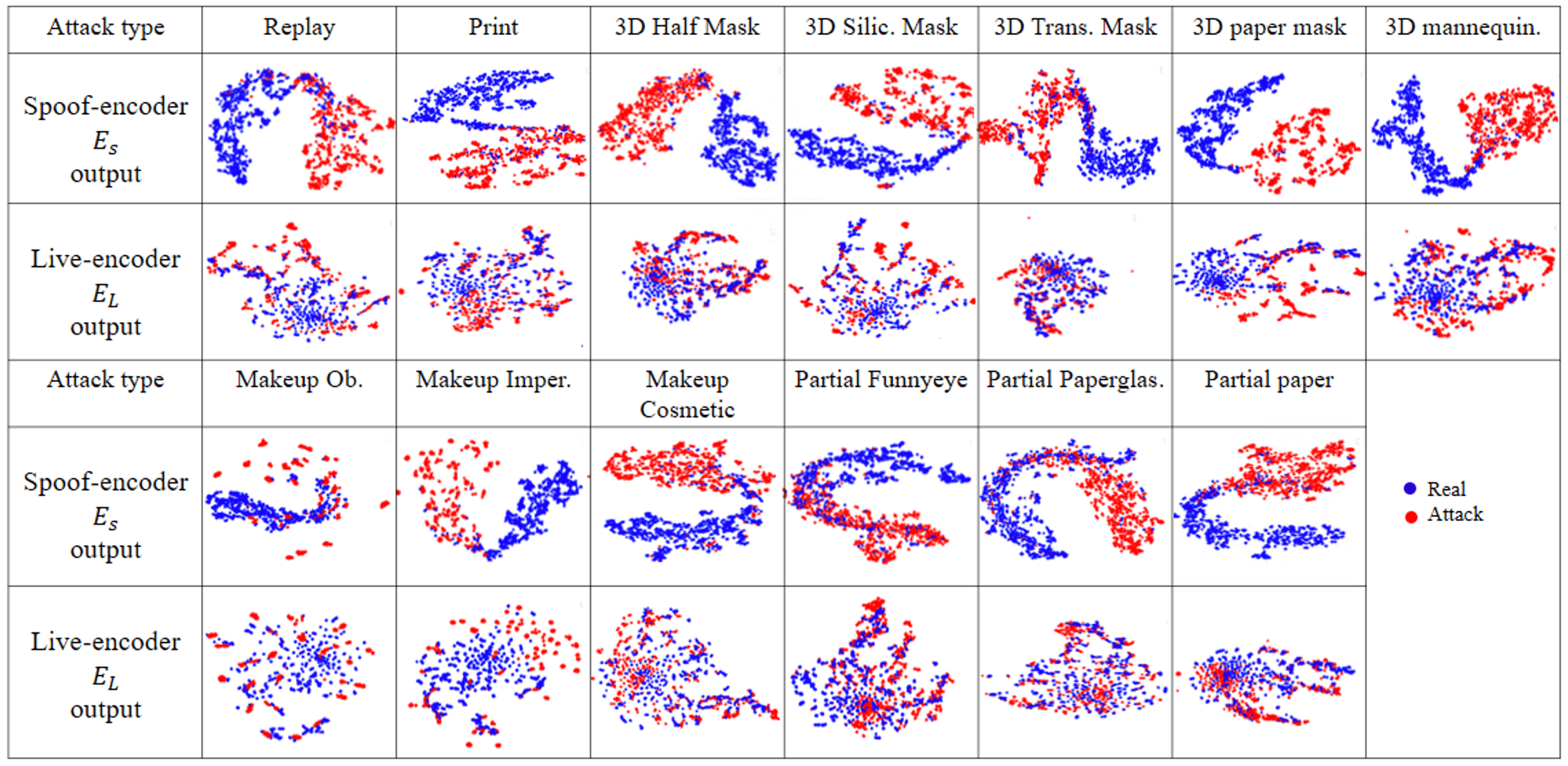}
   \caption{\textbf{Visualization of feature distributions of the SiW-M dataset by t-SNE\cite{van2008visualizing}. } We visualize the live feature $F_{L}$ and spoof feature $F_{S}$ in the SiW-M dataset of different unknown attack types. $F_{L}$ of real and attack samples are highly overlapped.  By contrast, $F_{S}$ distinguishing.
}
\label{fig2}
\end{figure*}

\subsection{Testing on SiW}

The SiW dataset contains 4,478 videos. It defines three sub-protocols to evaluate the model generalization ability under different face expressions, mediums, and attack types, respectively. The evaluation results are reported in Table \ref{siw}. Our method performs the best in sub-protocol 1 and sub-protocol 2. In SiW protocol 3, the training data only contains one type of attack, which could possibly inhibit the disentangled  feature  learning  in  our  network. We obtain competitive performance with 3.58 ACER $(\%)$. The experimental results show that our method has great generalization ability not only on the unseen attacks but also on different poses, attack mediums, and camera sensors. 

\begin{table}[ht]
\hspace*{-0.4cm}
\resizebox{1\linewidth}{!}{
\begin{tabular}{|c|c|c|c|c|}
\hline
Protocol           & Method & APCER$(\%)$  & BPCER$(\%)$  & ACER$(\%)$   \\ \hline
\multirow{5}{*}{1} & \begin{tabular}[c]{@{}c@{}}HMP \cite{yu2020face}\\ \end{tabular}    & 0.55       & 0.17       & 0.36       \\ \cline{2-5} 
                   & \begin{tabular}[c]{@{}c@{}}DRL \cite{zhang2020face}\\ \end{tabular}    & 0.07       & 0.5        & 0.28       \\ \cline{2-5} 
                   &\begin{tabular}[c]{@{}c@{}}DST \cite{liu2020disentangling}\\ \end{tabular}     & 0.00          & 0 .00         & \textbf{0.00}          \\ \cline{2-5} 
                   & LGC \cite{feng2020learning}    & 0.00          & 0.5        & 0.25       \\ \cline{2-5} 
                   & \textbf{Ours}   & 0.00          & 0.00          & \textbf{0.00}          \\ \hline
\multirow{5}{*}{2} & \begin{tabular}[c]{@{}c@{}}HMP \cite{yu2020face}\\ \end{tabular}     & 0.08$\pm$0.17 & 0.15$\pm$0.00 & 0.11$\pm$0.08 \\ \cline{2-5} 
                   & \begin{tabular}[c]{@{}c@{}}DRL \cite{zhang2020face}\\ \end{tabular}    & 0.08$\pm$0.17 & 0.21$\pm$0.16 & 0.15$\pm$0.14 \\ \cline{2-5} 
                   &\begin{tabular}[c]{@{}c@{}}DST \cite{liu2020disentangling}\\ \end{tabular}     & 0.00$\pm$0.00 & 0.00$\pm$0.00 & \textbf{0.00$\pm$0.00} \\ \cline{2-5} 
                   & LGC\cite{feng2020learning}    & 0.00$\pm$0.00 & 0.00$\pm$0.00 & \textbf{0.00$\pm$0.00} \\ \cline{2-5} 
                   & \textbf{Ours}   & 0.00$\pm$0.00 & 0.00$\pm$0.00 & \textbf{0.00$\pm$0.00} \\ \hline
\multirow{5}{*}{3} & \begin{tabular}[c]{@{}c@{}}HMP \cite{yu2020face}\\ \end{tabular}     & 2.55$\pm$0.89 & 2.34$\pm$0.47 & 2.45$\pm$0.68 \\ \cline{2-5} 
                   & \begin{tabular}[c]{@{}c@{}}DRL \cite{zhang2020face}\\ \end{tabular}   & 9.35$\pm$6.14 & 1.84$\pm$2.60 & 5.59$\pm$4.37 \\ \cline{2-5} 
                   &\begin{tabular}[c]{@{}c@{}}DST \cite{liu2020disentangling}\\ \end{tabular}     & 8.3$\pm$3.3   & 7.3$\pm$3.3   & 7.9$\pm$3.3   \\ \cline{2-5} 
                   & LGC \cite{feng2020learning}    & 1.61$\pm$1.69 & 0.77$\pm$1.09 & \textbf{1.19$\pm$1.39 }\\ \cline{2-5} 
                   & \textbf{Ours}   & 4.77$\pm$5.04 & 2.44$\pm$2.74 & 3.58$\pm$3.93 \\ \hline
\end{tabular}}
\caption{Comparison of state-of-the-art methods on the three SiW protocols \cite{liu2018learning}.}
\label{siw}
\end{table}

\begin{figure*}[ht]
\centering
   \includegraphics[width=1\linewidth]{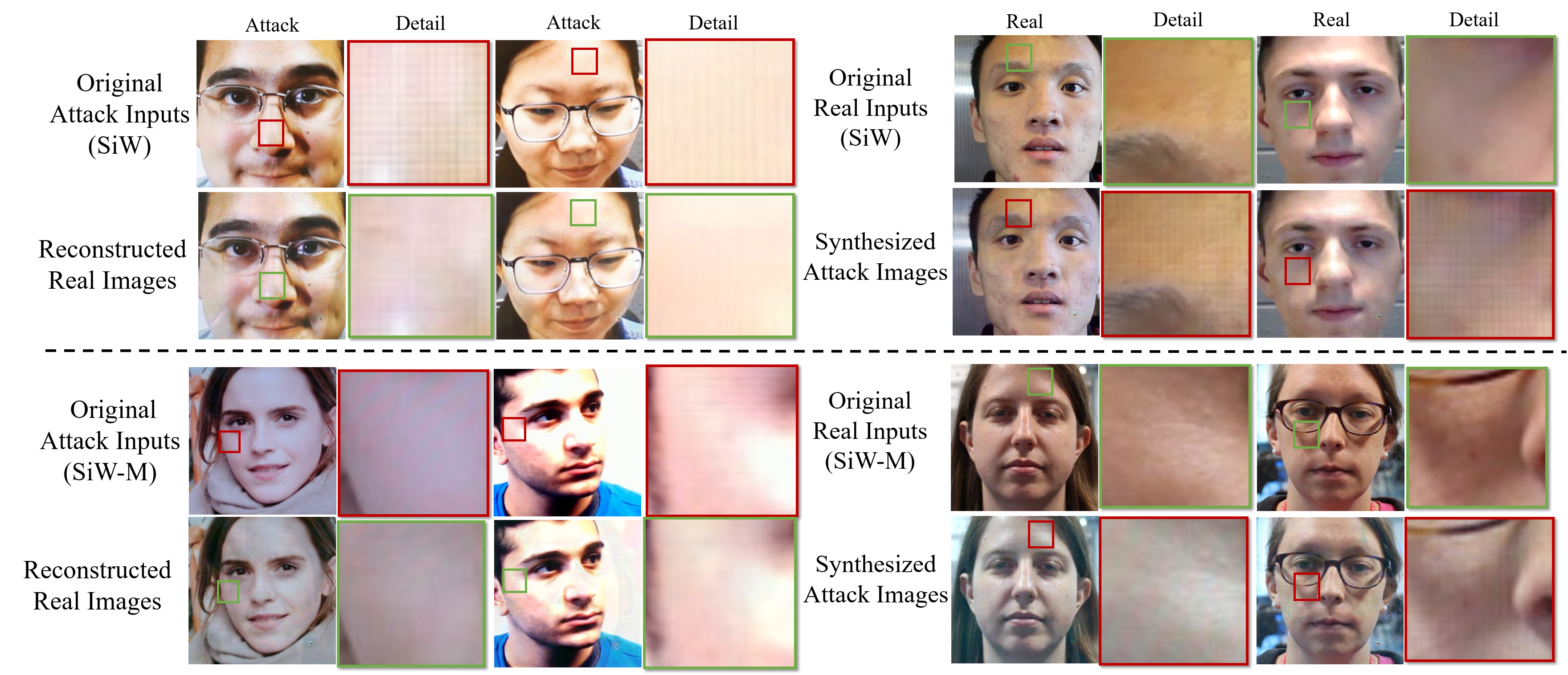}
   \caption{\textbf{Illustrations of the translation result in the SiW and the SiW-M datasets.} We swap the spoof features between a real person and a replay attack image to synthesize new translation images. The detailed images show that the screen textures have an apparent difference between the original inputs and the translation images.}
\label{fig1}
\end{figure*}

\section{Ablation Study}

\subsection{Discussion of the disentangled features}

Our disentanglement module is designed to disentangle between spoof features and live features. To verify the rationality of the disentangled features, we swap the spoof features between a real image and an attack image to synthesize a new translation image in this section.  
We choose two unpaired images A and B from the SiW dataset and the SiW-M dataset, where A is a real image, and B is an attack image. We experiment on the replay attack type because their screen textures can help us observe the discrepancy between real and attack images.

We use Live-encoder $E_{L}$ and Spoof-encoder $E_{S}$ to encode inputs into live features and spoof features, respectively. We obtain $F_{L}^{A}$, $F_{S}^{A}$, $F_{L}^{B}$, and $F_{S}^{B}$ after we encode images A and B. We exchange the spoof feature $F_{S}$ of A and B. Then, we use the decoder $D_{syn}$ to synthesize new real data $B^T$ and attack data $A^T$. This translation procedure can be described as:
\begin{equation}
\begin{split}
&F_{L}^{A} = E_{L}(A), \ F_{S}^{A} = E_{S}(A)\\
&F_{L}^{B} = E_{L}(B), \ F_{S}^{B} = E_{S}(B)\\
&A^T = D_{syn}(F_{L}^{A},F_{S}^{B}), \
B^T = D_{syn}(F_{L}^{B},F_{S}^{A})
\end{split}
\end{equation}

We show the translation results of some samples from SiW and SiW-M datasets in Figure \ref{fig1}. Images in the first and the third row are original inputs. The second and the fourth row depicts the generated outputs by the above translation process. We use red boxes to point out the details of the attack images, and green boxes refer to the details of the real images. As shown in the figure, we can easily observe the notable local detailed difference between original images and translation results. The original attack images are usually with repetitive striped patterns. After the translation process, we can notice that the translated images have fewer streaks on their surfaces. We also conduct the translation process for the original real data and generate the new synthetic attack images. We can notice that these new synthetic attack images contain some repetitive striped patterns. According to the translation results, we can confirm that our disentangled spoof features are critical for generating spoof images. With these spoof features, we can further employ these spoof information to distinguish attack data. 

\begin{table*}[ht]
\resizebox{1\textwidth}{!}{
\begin{tabular}{c|c|c|c|c|c|c|c|c|c|c}
\hline
\multirow{2}{*}{Method}        & \multicolumn{3}{c|}{CASIA-MFSD} & \multicolumn{3}{c|}{Replay-attack} & \multicolumn{3}{c|}{MSU-MFSD}     & \multirow{2}{*}{Overall} \\ \cline{2-10}
                               & Video   & Cut photo  & Wrapped  & Video  & Digital Photo  & Printed  & Printed & HR video & Mobile Video &            
                                             \\ \hline\hline
Proposed w/o first-stage     & 96.25   & 96.88      & 98.8     & 84.53  & 99.76           & 100      & 83.21   & 98.57      & 97.10        & 95.01$\pm$6.45               \\ \hline
Proposed w/o discriminator     & 95.36   & 96.97      & 98.5     & 82.59  & 99.7           & 100      & 76.57   & 100      & 96.35        & 94.0$\pm$8.47               \\ \hline
Proposed w/o aux classifier    & 96.69   & 98.47      & 97.77    & 93.65  & 99.23          & 99.92    & 76.5    & 82.07    & 95.42        & 93.3$\pm$8.29               \\ \hline\hline
Proposed w/o triplet loss      & 94.50   & 95.9       & 93.88    & 92.87  & 99.4           & 99.93    & 82.07   & 99.0     & 97.54        & 95.01$\pm$5.48              \\ \hline
Proposed w normal triplet loss & 96.66   & 98.83      & 99.72    & 97.34  & 99.76          & 100      & 84.25   & 99.4     & 99.28        & 97.25$\pm$5.01              \\ \hline\hline
\textbf{Proposed (full method)}         & 98.05   & 99.36      & 99.83    & 99.59  & 99.96          & 100      & 86.78   & 100      & 98.39        & \textbf{98.0$\pm$4.26}               \\ \hline\hline
\end{tabular}}
\caption{\textbf{The ablation study of the proposed method.} Experimental results of removing some components from the proposed system with ‘leave one attack type out’ protocol in the  CASIA-MFSD\cite{zhang2012face} dataset, the Replay-Attack\cite{chingovska2012effectiveness} dataset, and the MSU-MFSD\cite{zhang2012face} dataset.} 
\label{ablation}
\end{table*}

\subsection{Contribution of each Loss function}

In this section, we train the network with some components disabled or some loss excluded to compare the corresponding accuracies of the model. As shown in Table \ref{ablation}. The first-row method means that we train the method with the weights of the encoder $E_{L}$ randomly initialized rather than use the pre-trained encoder $E_{L}$ in the first stage.By comparing the results, we can see that $w/o$ first stage has much lower accuracy in the experimental results. Our two-stage training can enhance the model performance by extracting more refined disentangled spoof features.

Then we train the model without using adversarial loss. We can note that the performance drops significantly in some attack types; it may be because the decoder is required to reconstruct the images only by $L_{recon}$ loss. And then, we remove the $C_{aux}$ classifier. Without using binary supervision, the model is likely to overfit to the training set.

We also analyze the effectiveness of our loss $L_{t}$. One is only using normal triplet loss, and the other is abandoning the loss $L_{t}$. The poor accuracy shows that triplet learning can help the model decode more discriminative spoof maps. 
Using combinations of hard and normal triplet loss achieves better performance than merely using normal triplet. It indicates that the combinations of hard and normal triplet loss can facilitate contracting real samples and separating real samples and attack samples in the spoof feature space. Eventually, the proposed model training with all loss functions performs the best.

\subsection{Visualization and Analysis}
We visualize the feature embeddings $F_{L}$ and $F_{S}$ of our Live-encoder $E_{L}$ and Spoof-encoder $E_{S}$ with t-SNE \cite{van2008visualizing} to justify our disentanglement representation strategy. We adopt the leave-one-out method on the SiW-M dataset of different unknown attack types. We randomly choose 1000 real data and 1000 unseen spoof-type data from the testing set.
The live features represent the spoof-irrelevant feature in both attack samples and real samples like background and identification; hence, the live feature distribution of real data and attack data overlap at a high degree. On the other hand, our Spoof encoder $E_{S}$ focuses on extracting the discrepancy between attacks and real samples. Therefore, the spoof feature visualization result of real data and attack data are more separated. The input images of the Spoof-encoder $E_{S}$ and Live-encoder $E_{L}$ are the same, but their visualization results are highly dissimilar. The difference between their visualization results demonstrates that our method can extract disentangled features. Additionally, we show that the feature embeddings $F_{L}$ and $F_{S}$ are disentangled in the supplementary material.

\begin{figure}[ht]
    \centering
    \hspace*{-1.2cm}
    \includegraphics[width=1.2\linewidth]{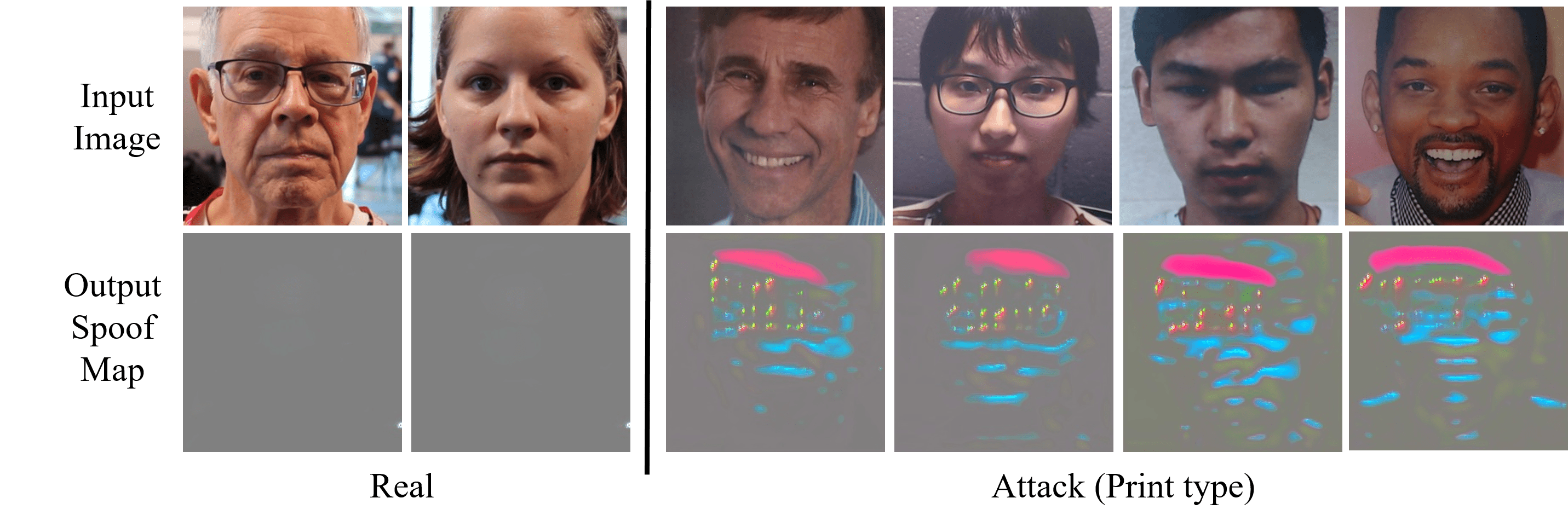}
    \caption{Examples of testing data in the SiW-M dataset and the corresponding spoof maps obtained by the proposed network. }
    \label{fig3}
    \vspace{-6.0mm}
\end{figure}

We also visualize the output of the decoder $D_{map}$ to describe how we detect spoof attack. We test the model with unseen attack types, and the generated spoof maps are depicted in Figure \ref{fig3}. The first row shows the input images, and the second row depicts the generated spoof maps.
As shown in Figure \ref{fig3}, the spoof maps for real face images are nearly zero maps, and there are some high-value attention positions in the spoof maps for attack samples.

\section{Conclusion}
We proposed a disentangled representation Network with dual-stage feature learning for face anti-spoofing. Unlike previous end-to-end architecture works, the dual-stage training design can improve the training stability and effectively encode the features to discriminate between real and attack. We further learn unsupervised spoof maps through our disentangled spoof features without using auxiliary information. These spoof features demonstrate great generalization ability when facing attacks of unknown spoof types.  Extensive experiments on public FAS datasets demonstrate that the proposed method achieves state-of-the-art performance for unseen spoof types FAS tasks. Our method has promising performance in unknown attack protocols. However, there is some room to progress when the training data only contains few attack types like the SiW dataset sub-protocol 3. In the future, we plan to extend our dual-stage method to remove more spoof-irrelevant factors, further improving our model generalization ability.

{\small
\bibliographystyle{ieee_fullname}
\bibliography{egbib}
}

\end{document}